\documentclass[lettersize,journal]{IEEEtran}
\usepackage[utf8]{inputenc}
\usepackage[T1]{fontenc}
\usepackage{amsmath,amsfonts,amssymb}
\usepackage{algorithmic}
\usepackage{algorithm}
\usepackage[caption=false,font=normalsize,labelfont=sf,textfont=sf]{subfig}
\usepackage{stfloats}
\usepackage{url}
\usepackage{verbatim}
\usepackage{graphicx}

\usepackage{cite}
\usepackage{multirow}
\usepackage{makecell}
\usepackage{balance}
\usepackage[section]{placeins}
\usepackage{tipa}
\usepackage{xcolor}
\usepackage{svg} 
\usepackage{pdfpages}
\usepackage{tabularx}
\usepackage{hyperref}
\usepackage{booktabs}
\usepackage{nicefrac}
\usepackage{microtype}
\usepackage{wrapfig}
\usepackage{enumitem}
\usepackage[normalem]{ulem}
\usepackage{colortbl}
\usepackage{newunicodechar}
\usepackage{changepage}
\newunicodechar{ˈ}{"}
\newunicodechar{ə}{@}
\newunicodechar{ɛ}{E}
\newunicodechar{ɪ}{I}
\newunicodechar{ˌ}{""}
\newunicodechar{ː}{:}
\newunicodechar{ɜ}{3}
\definecolor{LightCyan}{rgb}{0.88,1,1}
\hypersetup{
    colorlinks=true,
}
\usepackage{array}
\begin{document}
\title{Delayed Memory Unit: Modelling Temporal Dependency Through Delay Gate \\}

\author{Pengfei Sun, Jibin Wu$^{*}$, Malu Zhang, Paul Devos, and Dick Botteldooren
\thanks{This work was supported in part by the Research Foundation - Flanders under grant number G0A0220N and the Flemish Government under the "Onderzoeksprogramma Artificiele Intelligentie (AI) Vlaanderen", National Natural Science Foundation of China (Grant No. 62306259 and 62106038), Research Grants Council of the Hong Kong SAR (Grant No. C5052-23G and PolyU25216423), and the Sichuan Science and Technology Program under Grant 2023YFG0259.}

\thanks{Pengfei Sun, Paul Devos, and Dick Botteldooren are with the Department of Information Technology, WAVES Research Group, Ghent University, Technologiepark Zwijnaarde 126, 9052 Ghent, Belgium (e-mail: pengfei.sun@ugent.be; p.devos@ugent.be;  dick.botteldooren@ugent.be).

Jibin Wu is with the Department of Data Science and Artificial Intelligence and the Department of Computing, The Hong Kong Polytechnic University, Hong Kong SAR. ($^{*}$Corresponding Author${:}$ jibin.wu@polyu.edu.hk)

Malu Zhang is with the School of Computer Science and Engineering,
University of Electronic Science and Technology of China, Chengdu
610054, China (e-mail: maluzhang@uestc.edu.cn)
}}

\maketitle

\begin{abstract}
Recurrent Neural Networks (RNNs) are widely recognized for their proficiency in modeling temporal dependencies, making them highly prevalent in sequential data processing applications.  Nevertheless, vanilla RNNs are confronted with the well-known issue of gradient vanishing and exploding, posing a significant challenge for learning and establishing long-range dependencies. Additionally, gated RNNs tend to be over-parameterized, resulting in poor computational efficiency and network generalization. To address these challenges, this paper proposes a novel Delayed Memory Unit (DMU). The DMU incorporates a delay line structure along with delay gates into vanilla RNN, thereby enhancing temporal interaction and facilitating temporal credit assignment. Specifically, the DMU is designed to directly distribute the input information to the optimal time instant in the future, rather than aggregating and redistributing it over time through intricate network dynamics. Our proposed DMU demonstrates superior temporal modeling capabilities across a broad range of sequential modeling tasks, utilizing considerably fewer parameters than other state-of-the-art gated RNN models in applications such as speech recognition, radar gesture recognition, ECG waveform segmentation, and permuted sequential image classification.
\end{abstract}

\begin{IEEEkeywords}
recurrent neural network, delay gate, delay line, speech recognition, time series analysis
\end{IEEEkeywords}

\section{INTRODUCTION}
\IEEEPARstart{R}{ecurrent} neural networks (RNNs) have demonstrated remarkable performance in processing sequential data. Notable applications include speech recognition \cite{graves2013speech, sak2014long}, gesture recognition \cite{murakami1991gesture,lstmgesture}, and time series analysis \cite{Markovian}. However, despite being equipped with advanced optimization algorithms, vanilla RNN remains susceptible to vanishing and exploding gradient problems \cite{bengio1994learning}. These issues hinder their ability to learn and establish long-range temporal dependencies. 
Specifically,  when gradients vanish during backpropagation, small changes to the weights have a minimal effect on distant future states. On the other hand,  when gradients explode, gradient-based optimization algorithms encounter challenges in smoothly navigating the loss surface \cite{pascanu2013difficulty}. 

To address these issues, numerous novel neural architectures and training methods have emerged in recent decades. Notably, seminal neural architectures like the long short-term memory (LSTM) \cite{hochreiter1997long}, have introduced memory cells along with various gating mechanisms to facilitate the retention of historical information over extended periods. These gating mechanisms control information updates to the memory cells, thereby preventing rapid gradient vanishing or exploding over time. However, these gated RNNs often suffer from over-parameterization. For instance, in some speech processing tasks, it has been observed that both the update and reset gates exhibit similar behaviors, resulting in a large number of redundant parameters \cite{vanelli2018light}. Moreover, within the LSTM framework, the presence of three gates alongside memory cells leads to increased model evaluation time \cite{faraji2023new}. 
These challenges pose significant obstacles in terms of computational efficiency and may potentially give rise to overfitting issues. Consequently, ongoing research endeavors have been dedicated to developing alternative architectural solutions. One prominent example is the Gated Recurrent Unit (GRU), which reduces the number of gating units in LSTM to two~\cite{chung2014empirical}. 
However, it is crucial to acknowledge that despite these advancements, some degree of redundancy in parameterization still exists within these gating mechanisms \cite{ravanelli2017improving}.

The Transformer, a model introduced by Vaswani et al. \cite{vaswani2017attention}, has firmly established itself as a powerful solution for sequence modeling. Its success can be largely attributed to the incorporation of the self-attention mechanism, which excels in establishing temporal dependencies and facilitating temporal parallelization. While Transformers have demonstrated impressive performance across a range of sequential tasks, recent studies have highlighted potential challenges they may face when applied to smaller datasets or deployed on low-power devices~\cite{zheng2020improving,ezen2020comparison,deletang2022neural}. Unlike Transformers, RNNs offer the advantage of parameter sharing, enabling them to flexibly handle sequences of varying lengths and facilitating efficient deployment in practical scenarios. 

Recently, there has been a growing interest in incorporating neuronal delays into neural networks to enhance their temporal modeling capabilities. Studies on biologically plausible spiking neural networks (SNNs) suggest that the integration of delay line structures, represented by learnable synaptic or axonal delay variables, can significantly improve the temporal modeling capability of SNNs~\cite{hussain2014delay,taherkhani2015dl,shrestha2018slayer,zhang2020supervised,sun2022axonal,sun2023adaptive,sun2023learnable, hammouamri2023learning}. 
Moreover, research demonstrates that employing delay lines with a fixed number of delays proves effective for tasks such as sound source localization~\cite{jeffress1948place,carr1988axonal,4359176,pan2021multi}. In these models, neurons equipped with different delay parameters are utilized for coincidence detection, allowing for the detection of interaural time differences (ITD) between spatially distributed microphone pairs. While these delay-line models excel in feature extraction, their potential impact on the network dynamics of RNNs remains unclear.

In this paper, we address this research question by exploring the interplay between the delay line and RNN, which leads to the development of a novel RNN model, referred to as the Delayed Memory Unit (DMU). The DMU can effectively facilitate temporal information interaction and alleviate the challenge of learning long-range temporal dependencies within conventional RNNs.  Different from other gated RNNs that employ parameterized memory units to store historical information, DMU utilizes delay lines to directly propagate information into future time steps. This not only reduces the model parameters but also facilitates temporal information processing. Notably, converting a vanilla RNN to the proposed DMU is straightforward, requiring only the addition of delay gates. We conducted extensive experiments on a range of benchmark tasks, including audio processing, radar gesture recognition, waveform segmentation, and sequence classification. Notably, DMU surpassed other state-of-the-art (SOTA) gated RNNs, achieving superior performance with significantly reduced parameters. We also performed ablation studies to understand the impact of various hyperparameters associated with the delay gate. To summarise, our contributions are threefold:
\begin{itemize}

  \item {We proposed a novel RNN model for sequential modeling, dubbed DMU. This model integrates a delay line with a vanilla RNN to enhance temporal modeling capabilities. Theoretical analysis confirms the effectiveness of DMU in tackling the long-range temporal credit assignment problem.
  }
  \item {We have introduced two methodologies to reduce the computational cost of the DMU: integrating dilated delay and implementing thresholding schemes. These methodologies promise to enhance computational efficiency while preserving the commendable performance of DMU.
  }
  \item {We demonstrate the superior temporal modeling performance of DMU across a range of benchmark tasks, consistently outperforming state-of-the-art RNN models while utilizing significantly fewer parameters. 
  }

\end{itemize}
The rest of this paper is organized as follows. In Section \ref{method}, we first introduce the formulation of the proposed DMU, followed by an in-depth analysis of its effectiveness in facilitating temporal credit assignment. Furthermore, we conduct an analysis on the model complexity and provide two methods to enhance the computational efficiency of DMU. In Section \ref{experiment}, we evaluate the proposed DMU across a diverse range of temporal processing tasks. Furthermore, in Section \ref{Analysis}, we conduct a comprehensive study on the effectiveness of the proposed DMU model as well as its associated hyperparameters. Finally, we conclude the paper in Section \ref{conclusion}.
 
\section{Method}
\label{method}
\subsection{Delayed Memory Unit}
Given an input \( x_t \in \mathbb{R}^M \) and the  hidden state \( h_{t-1} \in \mathbb{R}^N \) at time step \( t-1 \), the hidden state at time step \( t \) can be updated as: 
\begin{equation}
  \tilde{h}_t = \sigma_g(W_{h}x_t + U_{h}h_{t-1} + b_{h}),
\end{equation}
where \( \sigma_{g} \) denotes a nonlinear activation function. The matrices \( W_{h} \in \mathbb{R}^{N \times M} \) and \( U_{h} \in \mathbb{R}^{N \times N} \) represent the feedforward and recurrent weights, respectively, while \( b_{h} \in \mathbb{R}^{N} \) is the bias term associated with the hidden neurons.

The key feature of our proposed DMU is the incorporation of the delay gate \( d_t \), which serves the purpose of dynamically controlling the propagation of the hidden state \( \tilde{h}_t \) to future time steps. This gating mechanism is mathematically expressed as follows:
\begin{equation}
d_t = \sigma_{h}( W_{d}x_t + U_{d}h_{t-1}^d + b_{d}),
\end{equation}
where \( W_{d} \in \mathbb{R}^{n \times M} \) and \( U_{d} \in \mathbb{R}^{n \times n} \) are the parameters governing the delay gate \( d_t \).  Here, \( n \)
represents the number of delays within the delay line. The value of \( n \) can be adjusted according to the specific characteristic of the task at hand, and setting \( n=0 \) will transform the DMU to the conventional RNN. To ensure that the output of the delay gates falls within a proper range, the softmax activation function \( \sigma_{h} \) is applied. 
The hidden state for the delay gate, denoted as \( h_{t-1}^d \), is updated in a similar manner to the hidden state \( \tilde{h}_t \). This state serves two purposes. Firstly, it helps to alleviate the workload on \( \tilde{h}_t \), allowing it to primarily store the information relevant to the task. Secondly, it introduces momentum and prevents abrupt changes in the delay gate outputs, thus enhancing noise robustness.

\begin{figure*}[tb]
\centering

{\includegraphics[width = 1\linewidth]{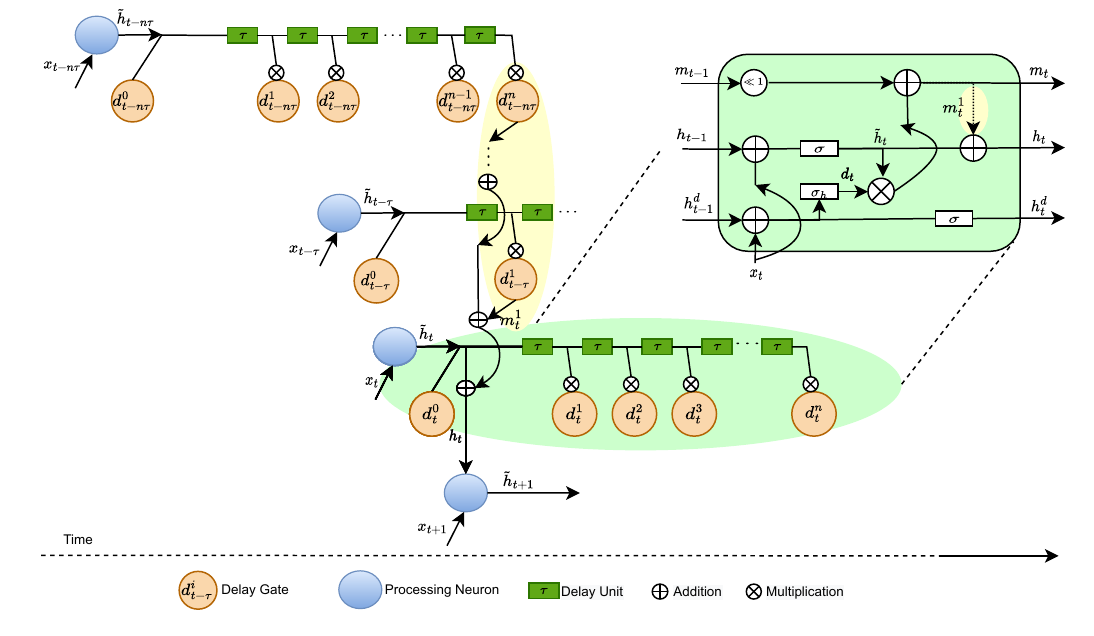}}
\caption{Illustration of the proposed DMU. The hidden state passes through a delay line, along which the delay unit continuously applies a fixed delay of \( \tau \) to the signal. At each point along the line, the corresponding delay cell gates the information from the state. The light blue area on the right side illustrates the time-unrolled DMU internal operation. This area corresponds to another light blue section at the bottom of the illustration. Similarly, the segment highlighted in light yellow within the figure represents the first state of the sliding memory window, $m_t^1$. }
\label{sliding_delaygate}
\vspace{0.2cm}
\end{figure*}

As shown in Fig. \ref{sliding_delaygate}, the neuron at time \( t \) receives information not only from the previous hidden state \(  h_{t-1} \) but also from the delayed information controlled by the corresponding delay gates. This can be represented by: 
\begin{equation}
\label{last45qww}
h_t = \tilde{h}_t + \sum_{i=t-n\tau}^{t-\tau} d^{(t-i)/\tau}_{i} *  \tilde h_{i},
\end{equation}
where \( d_i^{(t-i)/\tau} \) is the delay gate output at the delay line index \( (t-i)/\tau \) for time step \( i \). The \( \tau \) is the delay line dilation/skipping factor that represents the temporal resolution of the delay line. 

For efficient implementation of the proposed delayed gate, we employ a sliding window memory \( m_t \in \mathbb{R}^{N \times n} \) to store the delayed information. This memory is updated at each time step as: 
\begin{equation}
 m_{t} = \tilde h_t \otimes d_t   + {m_{t-1}{\ll}1},
\end{equation}
where \( \otimes \) represents the Kronecker product, and \( {\ll}1 \) denotes the operation of shifting the contents within the sliding window one step forward in time. Following this implementation, Eq. \ref{last45qww} can be updated according to:
\begin{equation}
    h_t = \tilde h_t + m^1_t, 
\end{equation}
where \( m^1_t \) represents the information that is to be incorporated into the current time step $t$. Specifically, \( m^1_t \) equals the sum of the delayed information, which is described by the second term of Eq. \ref{last45qww}. This formulation allows the current network state to be effectively influenced by the previous states ranging from time step \( t-n\tau \) to \( t-\tau \). Unlike the vanilla RNN and other gated variants, the DMU enables a more direct influence of historical network states on future states. This feature significantly facilitates temporal credit assignment, as elaborated in the subsequent section.

It is worth noting that the delay gate mechanism proposed in this work exhibits similarities to the skip connections originally introduced for image recognition tasks \cite{he2016deep}. In essence, both designs address the vanishing gradient problem by establishing shortcut connections that aid in the propagation of information. However, it is important to note that while skip connections in image recognition primarily focus on information propagation across the spatial dimension, the delay gate mechanism is specifically designed to address the temporal dimension.

\subsection{Temporal Credit Assignment within DMU}
\label{bptt}

Consider \( h_t \) and \( h_T \) as the hidden unit vectors at time steps \( t \) and \( T \), respectively, where \( t \ll T \). Let \( L \) be the loss function that we aim to minimize at time \( T \). The error gradient backpropagated from \( T \) to \( t \) can be expressed as:
\begin{equation}
\label{loss}
\begin{aligned}
     \frac{\partial L}{\partial h_t}  &= \frac{\partial L}{\partial h_T} \frac{\partial h_T}{\partial h_t} \\
    &= \frac{\partial L}{\partial h_T}{\frac{\partial (\tilde{h}_{T} + \sum_{i=T-n\tau}^{T-\tau} d^{(T-i)/\tau}_{i} \tilde h_{i})}{\partial h_t}} \\
    &= \frac{\partial L}{\partial h_T} \left( \frac{\partial \tilde{h}_{T}}{\partial h_t} + \frac{\partial \sum_{i=T-n\tau}^{T-\tau} d^{(T-i)/\tau}_{i} \tilde h_{i}}{\partial h_t} \right).
\end{aligned}
\end{equation}

To simplify our analysis, let's disregard the term \( \frac{\partial L}{\partial h_T} \), as it is not recursively affected by time. The second term within the bracket can be expressed as follows:
\begin{equation}
\begin{aligned}
  \alpha_{T} =& \lambda_{T} \alpha_{T-1} + \sum_{i=T-n\tau}^{T-\tau} d^{(T-i)/\tau}_{i} \lambda_{i} \\
  =& \prod_{k=t+1}^{T} \lambda_{k} + \sum_{i=t+2}^{T} \left( (\prod_{j=i}^{T} \lambda_{j}) \sum_{l=j-n\tau-1}^{j-\tau-1} d^{(T-l)/\tau}_{l} \lambda_{l} \right) \\
  &+ \sum_{i=T-n\tau}^{T-\tau} d^{(T-i)/\tau}_{i} \lambda_{i},
\end{aligned}
\end{equation}
where $\alpha_{k} = \frac{\partial {h}_{k}}{\partial h_{t}}$ and $\lambda_{k} = \frac{\partial \tilde{h}_{k}}{\partial h_{k-1}}$. $\prod_{k=t+1}^{T} \lambda_{k} = \prod_{k=t+1}^{T} \text{diag}(\sigma^{'}(h_{k})) U_{h}^T$
represents the gradient of the vanilla RNN, and $\text{diag}(\sigma^{'}(h_{k+1}))$
is the Jacobian matrix of the pointwise activation function. From this equation, it's evident that the delay gate introduces an addition term that can facilitate the smoother propagation of error gradients across longer time spans. This mechanism effectively addresses the issue of vanishing gradients, which can hinder the training process of RNNs.

\begin{table*}[t] 
  \caption{The network update formulation and parameter count for vanilla RNNs, DMU, and LSTM models.}
  \label{dmu}
    \centering
    
     \begin{minipage}[h]{0.48\textwidth} 
         \centering
        \begin{tabular}{l|l}\hline  
 \textbf{Formulation} & \textbf{Parameters} \\

\hline  
\multicolumn{1}{c|}{RNN} \\   
$h_t = \sigma_{g}(W_{h}x_t + U_{h}h_{t-1} + b_{h})$ & $N^2 +MN +N$\\  

\hline
\multicolumn{1}{c|}{DMU} \\  
$\tilde h_t = \sigma_{g}(W_{h}x_t + U_{h}h_{t-1} + b_{h})$  & $N^2+MN+N+Mn$\\
$d_t = \sigma_{h}( W_{d}x_t + U_{d}h_{t-1}^d  + b_{d})$ &$+n^2+n$   \\
$h_{t}^d = \sigma_{g}( W_{d}x_t +U_{d}h_{t-1}^d+b_d) $\\
$ m_{t} = d_t \otimes \tilde h_t + {m_{t-1}\ll 1} $  \\
$h_t = \tilde h_t + m^1_t$\\
\hline
\multicolumn{1}{c|}{LSTM} \\  
$i_t = \sigma_{g}(W_{i}x_t + U_{i}h_{t-1} + b_{i})$ & $4*(N^2+MN+N)$ \\
$f_t = \sigma_{g}(W_{f}x_t + U_{f}h_{t-1} +  b_{f})$\\
$o_t = \sigma_{g}(W_{o}x_t + U_{o}h_{t-1} +  b_{o})$ \\
$z_t = \tanh(W_{c}x_t + U_{c}h_{t-1} +  b_{c})$ \\
$c_t = f_t \circ c_{t-1} + i_t \circ z_t$ \\
$h_t = o_t \circ \tanh(c_t)$\\
\end{tabular}
 \end{minipage} 
\begin{minipage}[h]{0.48\textwidth} 
    \centering
\begin{tabular}{ll} 
\hline
\textbf{Notations}& \textbf{Descriptions}\\
\hline
 $x_t \in \mathbb{R}^M$&\text{Input} \\  \hline
$\tilde{h}_t$, ${h}_t \in \mathbb{R}^N$&\text{The hidden state} \\  \hline
 $\tilde{h}_t^d \in \mathbb{R}^n$ & \text{The delayed hidden state} \\ \hline
 $m_t \in \mathbb{R}^{N \times n}$&\text{The sliding memory window} \\ \hline
 $d_t \in \mathbb{R}^n$ & \text{The delay gate}\\  \hline
 $n$ & \text{The delay line length} \\  \hline
  $\tau$ & \text{The dilation factor}\\  \hline
  $W \in \mathbb{R}^{N \times M}$& \text{Learnable feedforward weight}\\  \hline
  $U \in \mathbb{R}^{N \times N}$& \text{Learnable recurrent weight}\\\hline
    $b\in \mathbb{R}^{N }$& \text{Learnable bias}\\\hline
   $\otimes$ & \text{The kronecker product }\\ \hline
     $\circ$ & \text{The hadamard product}\\ \hline
   $\sigma_g(\cdot)$ & \text{The tanh activation function }\\ \hline
   $\sigma_h(\cdot)$ & \text{The softmax activation function}\\ \hline
    ${\ll}1$ & \text{The sliding operation}\\ 
  
\hline
\end{tabular}
    \end{minipage} 
  \end{table*}

\subsection{Model Complexity Analysis}
Here, we compare the model complexity of the proposed DMU against vanilla RNN and LSTM. As shown in Table \ref{dmu}, the parameter count for the vanilla RNN is  $N^2+MN+N$, while an LSTM unit has $4*(N^2+MN+N)$ parameters. On the other hand, the DMU introduces an additional $Mn+n^2+n$ parameters compared to the vanilla RNN. However, given that $n \ll M \ll N$, the DMU only adds a slightly higher number of parameters compared to the RNN unit, while still maintaining approximately  $4\times$ fewer parameters than the LSTM unit. Although our approach requires additional memory to retain information in the sliding window memory, we provide two solutions that can effectively address this challenge in the following section.

\subsection{Thresholding DMU and Dilated Delay}
\label{thresholdingdmu}
As given earlier, the memory usage for the sliding window memory $m_t$ is $N*n$, and it can be efficiently managed by reducing the length of the delay line $n$. To this end, we propose two strategies. Firstly, we introduce a thresholding mechanism to skip less significant delay gates, which can be described as follows:

\begin{equation}
 d_t = \begin{cases}
          d_t  &  \text{if } d_t \geq \theta \\
          0 & \text{if } d_t < \theta\\
          \end{cases},
\end{equation}
where the threshold $\theta$ is utilized to determine whether the output of the delay gate should be skipped. Specifically, any delay gate outputs falling below the value of $\theta$ will be considered insignificant, which can be safely ignored without affecting the task performance too much.

Additionally, as depicted by the second term of Eq.~\ref{last45qww}, dilated delay can be utilized to skip elements on the delay line at an interval of $\tau$. This method retains the same temporal duration covered by the delay line, while reducing the total memory consumption by a factor of $\tau$. The impact of these two strategies on memory consumption and task performance are analyzed in details in Sections \ref{effect_tau} and \ref{threshold_scheme}.

\section{Experimental Results}
\label{experiment}
In this section, we evaluate the performance of the proposed DMU on real-world temporal classification tasks. We demonstrate the superiority of the proposed DMU by comparing its efficacy against other SOTA baseline models.

\subsection{Experimental Setups and Training Configuration}
For our model evaluations, we selected benchmark tasks encompassing an inherent temporal dimension. These tasks span areas such as speech processing, gesture recognition, waveform classification, and more. We employed the Pytorch library for implementing all models, except for the TIMIT phoneme recognition task where the PyTorch-Kaldi library \cite{pytorch-kaldi} was used. Across all tasks, we set the nonlinear activation function $\sigma_g$ to be $tanh(\cdot)$. However, for the TIMIT phoneme recognition task, we followed the default Pytorch-Kaldi library configuration, which employs $relu(\cdot)$. We selected the $softmax(\cdot)$ function for $\sigma_h$ to ensure a proper probability distribution. For all experiments, the delay line resolution, denoted by $\tau$, was set to 1 time step.

All models were trained using the Adam optimizer \cite{kingma2014adam}, with each training batch consisting of 128 data samples. We used a constant learning rate of $0.001$ across 120 training epochs. We adopted the Kaiming weight initialization approach \cite{he2015delving} for both network weights and bias. The sliding window memory was initialized with zeros. Notably, benefiting from the enhanced temporal credit assignment facilitated by the delay gate, DMU sidesteps the necessity for other advanced optimization methods, such as weight normalization, gradient clipping, and weight regularization, etc.  All model training was conducted on Nvidia Geforce GTX 1080Ti GPUs, each equipped with 12 GB of memory.

\subsection{Speech Processing}
\begin{table*}[t]
\centering 
        \normalsize
	\centering
	\caption{Comparison of model performance across TIMIT, SHD, Hey Snips, SoLi, QTDB, and PSMNIST datasets. The terms \#Units/\#Params denote the number of hidden neurons in the recurrent layer and the total parameters of the network, respectively. PER stands for Phone Error Rate. For the PSMNIST task, we replicated the results of Lipschitz RNN, coRNN, and $\tau$-GRU using publicly available source codes.} 

 \label{tbl:results}	

\begin{tabular}{lr}
 \subfloat[TIMIT]{
    \label{tbl:timit}
    \begin{tabular}{lrr}
        \hline
        \textbf{Methods} & \textbf{\#Params} & \textbf{PER(\%)} \\
        \hline
        RNN$^*$ & 3.23M & 18.8\% \\
        GRU$^*$ & 7.51M & 18.7\% \\
        LSTM$^*$ & 9.66M & 17.6\% \\
        Bidirectional-RNN$^*$ & 5.20M & 17.7\% \\
        Bidirectional-GRU$^*$ & 14.03M & 16.4\% \\
        Bidirectional-LSTM$^*$ & 14.36M & 15.9\% \\
        \cellcolor{LightCyan}DMU & \cellcolor{LightCyan}4.57M & \cellcolor{LightCyan}17.5\% \\
        \cellcolor{LightCyan}\bf{Bidirectional-DMU} & \cellcolor{LightCyan}\bf 6.59M & \cellcolor{LightCyan}\bf 15.6\% \\
        \hline
        \multicolumn{3}{l}{\footnotesize{$*$ Reproduce through Pytorch-Kaldi\cite{pytorch-kaldi}}}\\
		
    \end{tabular}
    }
\hspace{0.2mm}
    
\subfloat[SHD]{
        \label{tbl:shd}
        \begin{tabular}{lrr}
            \hline
            \textbf{Methods} & \textbf{\#Params} & \textbf{Acc.} \\ \hline
            FSNN \cite{cramer2020heidelberg}$^*$ & 0.09M & $48.10\%$ \\
            RSNN\cite{cramer2020heidelberg}$^*$ & 1.79M & $83.20\%$ \\
            Adaption RSNN \cite{yin2020effective}$^*$ & 0.14M & $84.40\%$ \\
            Attention RSNN \cite{yao2021temporal}$^*$ & 0.19M & $90.02\%$ \\
            Bi-LSTM\cite{yin2021accurate} & 1.10M & $87.20\%$ \\
            LSTM & 0.56M & $79.90\%$ \\
            \cellcolor{LightCyan}\textbf{DMU} & \cellcolor{LightCyan}0.16M & \cellcolor{LightCyan}$90.81\%$ \\
            \cellcolor{LightCyan}\textbf{DMU} & \cellcolor{LightCyan}0.24M & \cellcolor{LightCyan}\textbf{91.48\%} \\ \hline
            	\multicolumn{3}{l}{\footnotesize{$*$  Spiking Neural Network}}\\
        \end{tabular}
    }\\
 
\subfloat[Hey Snips]{
    \label{table:wake}
    \begin{tabular}{lrr}
        \hline
        \textbf{Methods} & \textbf{\#Units/\#Params} & \textbf{Acc.} \\
        \hline
        Spiking CNN\cite{yilmaz2020deep} & -/583k & 95.06\% \\
        Vanilla RNN & 64/88k & 95.59\% \\
        LSTM & 64/347k & 95.64\% \\
        GRU & 64/261k & 95.64\% \\
        \cellcolor{LightCyan}\bf DMU & \cellcolor{LightCyan}64/113k & \cellcolor{LightCyan}\bf 95.98\% \\
        \cellcolor{LightCyan}\bf DMU & \cellcolor{LightCyan}185/346k & \cellcolor{LightCyan}\bf 96.29\% \\
        \hline
    \end{tabular}
}
\hspace{1.2mm}
\subfloat[SoLi]{
\label{tbl:soli}
\begin{tabular}{lrr}
		\cline{1-3}
		\multicolumn{1}{l}{\textbf{Methods}}&  
		\multicolumn{1}{c}{\textbf{\#Params}}&		\multicolumn{1}{c}{ \textbf{Acc.}}		
		\\ \hline 
		CNN Deep \cite{wang2016interacting}&  17.04M &  48.18\% \\
		CNN-LSTM \cite{wang2016interacting}&  2.52M  &  87.17\%  \\
  		SRNN \cite{yin2020effective} &  1.05M &  91.90\%  \\ 
		Vanilla RNN & 1.05M &90.38\% \\
		   LSTM &  3.42M
		& 92.62\%\\ 
	   \cellcolor{LightCyan}\bf DMU  &\cellcolor{LightCyan}1.10M&\cellcolor{LightCyan}\bf94.78\% \\
		   \hline 
	\end{tabular}}\\ 
 
\subfloat[QTDB]{
\label{table:resultecg}
\begin{tabular}{lrr}
    \hline
    \textbf{Methods} & \textbf{\#Unit/\#Params} & \textbf{Acc.} \\
    \hline 
    ECGNet\cite{abrishami2018p} & -/8.64$k$ & 81.09\% \\ 
    Vanilla RNN & 32/3.43$k$ & 90.56\% \\
    GRU & 32/9.96$k$ & 93.73\% \\
    LSTM & 32/13.22$k$ & 94.18\% \\ 
    \cellcolor{LightCyan}\bf DMU & \cellcolor{LightCyan}32/5.82$k$ & \cellcolor{LightCyan}94.48\% \\ 
    \cellcolor{LightCyan}\bf DMU & \cellcolor{LightCyan}52/12.33$k$ & \cellcolor{LightCyan}\bf 94.97\% \\
    \hline
\end{tabular}
}

\subfloat[PSMNIST]{
\label{tbl:psmnist}
\begin{tabular}{lrr}
    \hline
    \textbf{Methods} & \textbf{\#Unit/\#Params} & \textbf{Acc.} \\
    \hline 
    GRU \cite{chandar2019towards} & -/165$k$ & 92.39\% \\	      
    LSTM \cite{chandar2019towards} & 200/165$k$ & 89.86\% \\
    Lipschitz RNN\cite{erichson2020lipschitz} & 200/82$k$ & 96.40\% \\   
    coRNN\cite{rusch2020coupled} & 200/82$k$ & 96.06\% \\
    $\tau$-GRU\cite{erichson2022gated} & 200/164$k$ & \bf96.97\% \\
    \cellcolor{LightCyan}\bf DMU & \cellcolor{LightCyan}200/49$k$ & \cellcolor{LightCyan}96.39\% \\
    \hline
\end{tabular}
}
 \end{tabular}
\end{table*}

\subsubsection*{\bf Phoneme Recognition}
The TIMIT dataset \cite{garofolo1993timit} is a classical benchmark used for phoneme recognition tasks. In this dataset, each utterance contains 16-bit speech waveforms sampled at 16 kHz, along with time-aligned orthographic, phonetic, and word transcriptions. We extract 39-dim Mel-frequency cepstral coefficients (MFCCs) \cite{davis1980comparison} from the raw audio waveforms and use them as input to our neural network-based classifier. To facilitate comparison with other baseline models, we follow the network and training configurations provided by the Pytorch-Kaldi library\footnote{
Pytorch-Kaldi is publicly available at: \url{https://github.com/mravanelli/pytorch-kaldi}}. For the DMU model, we set the number of delays to $n=45$. 

As shown in Table \ref{tbl:results}\textbf{(a)}, the proposed DMU model achieves a superior PER compared to other strong baseline models. Specifically, with only an additional 1.39 M parameters introduced by the delay gate, the PER is reduced by 2.1\% compared to the vanilla bidirectional-RNN model. Furthermore, our DMU model surpasses LSTM and GRU models by 0.3\% and 0.8\%, respectively. These results underscore the effectiveness of the proposed delay gate in tackling the challenging speech processing task.

\subsubsection*{\bf Wake-word Detection}
\begin{figure}
  \begin{center}    \includegraphics[width=0.5\textwidth]{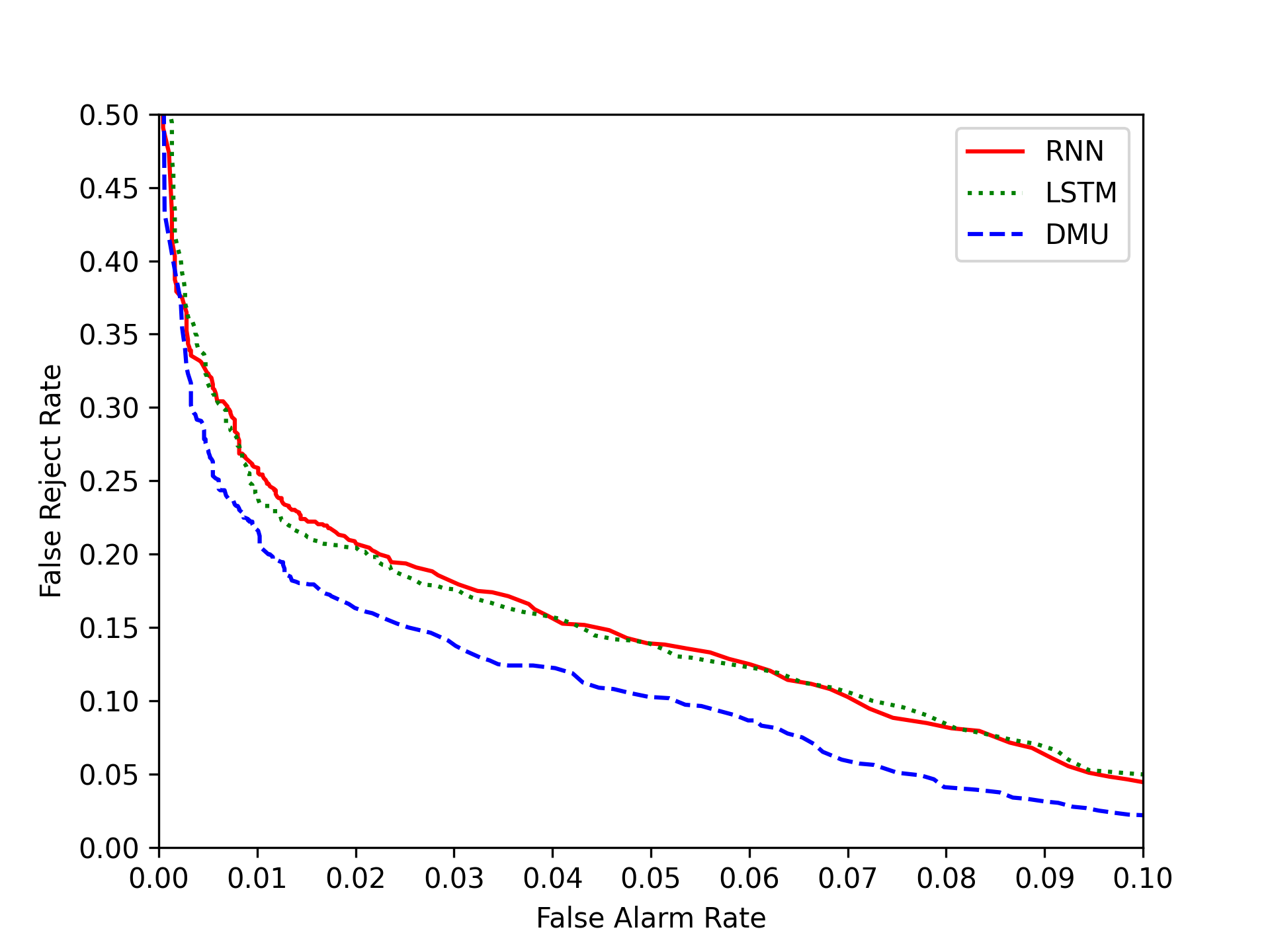}
  \end{center}
  \caption{Comparison of the false reject rate and false alarm rate between the proposed DMU model, RNN, and LSTM models on the wake-word detection task.}
  \label{kws}
  \vspace{0.2cm}
\end{figure}

The Hey Snips dataset \cite{coucke2019efficient} has been widely used for the wake-word detection task, which contains utterances collected from nearly 1,800 speakers. Following the experimental settings of Yılmaz et al. \cite{yilmaz2020deep}, we extract 40-dim MFCC features from raw audio waveforms, with a frame length of 30 $ms$ and a frame shift of 10 $ms$, resulting in a total of 98 frames for each 1-second input. For this task, we adopt a convolutional recurrent neural network (CRNN) architecture, where convolutional layers are introduced at the front-end to extract local spectrotemporal features and recurrent layers are further used to handle the global temporal transitions as described in \cite{yilmaz2020deep}. Our CRNN consists of one convolutional layer with 32 filters that have a kernel size of 5 and 10 for the frequency and time dimensions, respectively, followed by two recurrent layers with 64 neurons each. For the delay gate, the total number of delays $n$ is set to 20 due to the short duration of each sample. The other training configurations are consistent with those of Yılmaz et al. \cite{yilmaz2020deep}.

As presented in Table \ref{tbl:results}\textbf{(c)}, our results demonstrate that the proposed DMU consistently outperforms other baseline models when using the same number of neurons. Moreover, as shown in Fig. \ref{kws}, we notice that DMU exhibits higher robustness compared to RNN and LSTM models in terms of the False Rejection Rate (FRR) and False Alarm Rate (FAR).

\subsubsection*{\bf Event-based Spoken Word Recognition}

The Spiking Heidelberg Dataset (SHD) \cite{cramer2020heidelberg} was introduced to study event-based speech processing. It is constructed by converting raw audio waveforms into an event-based representation using a biologically plausible artificial cochlear model. In this study, we utilize the SHD dataset to evaluate the applicability and generalization of the DMU to other data modalities. Each sample in the SHD dataset comprises 700 channels, each corresponding to distinct frequency-sensitive neurons in the peripheral auditory system. To ensure sufficient temporal resolution while minimizing post-processing workload, we aggregate the spikes within each 10 $ms$ time bin, leading to a total of 100 time steps per sample. For this task, we employ a two-layer recurrent neural network and utilize the readout integrator \cite{yin2021accurate} for decoding, which has been shown to outperform the last time step decoding method \cite{cramer2020heidelberg}. The total number of delays $n$ is set to $30$ to encourage the DMU to learn short- and mid-range temporal dependencies between different phonetic units. 

The results of the DMU, along with other SOTA recurrent spiking neural networks (RSNNs) and LSTM models, are provided in Table \ref{tbl:results}\textbf{(b)}. Notably, our proposed DMU model has achieved the SOTA test results on this dataset, surpassing LSTM and Bi-LSTM models by 11.58\% and 4.28\%, respectively, while using fewer than half and a quarter of their respective network parameters. This result also competes favorably with the SOTA RSNN model that employs temporal attention and advanced data augmentation techniques \cite{yao2021temporal}. It's worth noting that when networks are trained with the last time step loss, all SNNs fail due to the requirement of long-range temporal credit assignment. Similarly, the LSTM can only achieve a test accuracy of $56.63\%$, while our DMU attains $78.71\%$.

\subsection{Radar Gesture Recognition}

Here, we perform a radio-frequency-based gesture recognition task using the SoLi dataset \cite{wang2016interacting}. In this dataset, a millimeter-wave radar captures reflected energy, mapping it to pixel intensity. We design the task to process radar samples sequentially, frame by frame, making a decision after the last frame has been read. Following the recommendation by Yin et al. \cite{yin2021accurate}, we employ a single channel instead of four, as it has been demonstrated to contain ample discriminative information \cite{yin2021accurate}. Our network architecture consists of a recurrent layer with 512 neurons, followed by a dense layer with 512 neurons, and a final classification layer with 11 neurons. To capture the desired temporal dependencies of this dataset effectively, we set the total number of delays to 40, ensuring an adequate range of temporal information integration.

As the test results reported in Table \ref{tbl:results}\textbf{(d)}, CNN models are inferior to both vanilla and gated RNN models in this context. This aligns with the insight that the radar sensor mainly captures hand pose changes, which convey more information compared to absolute positions \cite{lien2016soli}. Same as the previous tasks, the DMU outperforms the vanilla RNN by 4.4\%, with only a slight increase in the total parameter count. Additionally, it surpasses the performance of the LSTM model by $2.16\%$, utilizing less than a third of the LSTM model's parameters.

\subsection{ECG Waveform Segmentation}
  
The electrocardiogram (ECG) is a vital tool used by clinicians to evaluate a patient's cardiovascular system. For this task, we employ the QTDB dataset \cite{laguna1997database} to assess the predictive capability of our proposed DMU model. The objective is to segment the ECG signal into distinct characteristic waveforms (Normal, P, QR, RS, and T). Clinicians can then diagnose based on the shape and duration of these waveforms. The QTDB dataset comprises 105 ECG records, each with a duration of 15 minutes. With a sampling frequency of 250 Hz, we pad shorter signals with zeros to achieve a consistent length of 300 data points. Each recording in the QTDB dataset features two input channels. Inspired by the design of ECGNet \cite{abrishami2018p}, our network incorporates two recurrent layers, each containing 32 neurons. Considering the short- to mid-range temporal dependencies required in this task, we set the DMU delay to 30. As evidenced in Table \ref{tbl:results}\textbf{(e)}, the DMU consistently outperforms other competitive baseline models,  with similar or fewer parameters, showcasing its superior temporal modeling capability.

\subsection{Permuted Sequential Image Classification}
\label{psmnist}

\begin{figure}
  \begin{center}
    \includegraphics[width=0.5\textwidth]{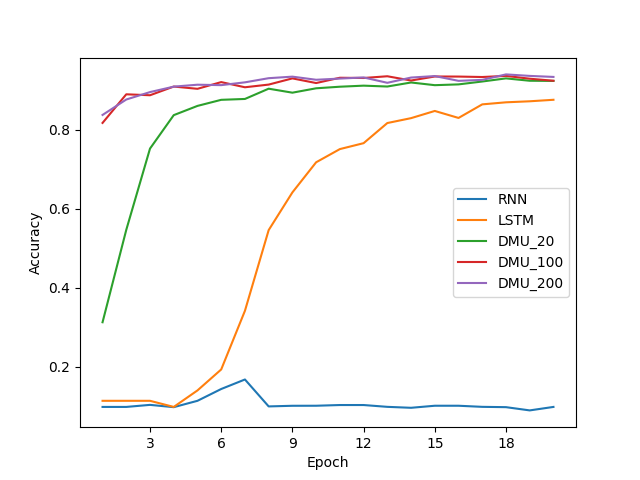}
  \end{center}
  
  \caption{{Comparison of the learning curves of different models on the PS-MNIST dataset with a temporal duration of 784 time steps. This dataset has been specifically designed to test the model's ability to retain long-term memory between pixels that may be widely separated.} All the models use $200$ hidden neurons, and DMU\_$n$ denotes the DMU model with the number of delays of $n$. }
  \label{psmnist_1}
\end{figure}

The permuted sequential MNIST (PS-MNIST) dataset\cite{le2015simple} has been designed to evaluate the capability of RNN models in modeling long-range temporal dependencies. This dataset is derived from the MNIST dataset by first flattening each image into a 784-dim vector, then shuffling its spatial information with a consistent permutation vector. In our experiments, the elements of this vector are sequentially presented to the network, with decisions made after all pixels have been processed. This task demands the model to decipher the original temporal order and capture dependencies between different pixels. Following the experimental setup of Chandar et al. \cite{chandar2019towards}, our DMU is designed with a single recurrent layer with the number of delays $n$ set to $80$. 

Results in Table \ref{tbl:results}\textbf{(f)} highlight that our proposed DMU can achieve a test accuracy of $96.39\%$ using only $49K$ parameters. This represents a substantial accuracy improvement of 6.53\% over LSTM, despite employing only 30\% of its parameters. Furthermore, our DMU is competitive with other recently introduced RNN models engineered for learning long-term dependencies, such as Lipschitz RNN, coRNN, $\tau$-GRU~\cite{rusch2021long, erichson2022gated, rusch2020coupled, erichson2020lipschitz}. Although the $\tau$-GRU demonstrates remarkable performance in this challenging task, it requires specialized training techniques to achieve the peak performance. We also provide the learning curves in Fig. \ref{psmnist_1} to compare the learning dynamics of our DMU with other baseline models. Notably, under identical training conditions and neuron count, the DMU converges rapidly within ten epochs, significantly surpassing the LSTM model.

\section{Discussions}
\label{Analysis}
In this section, we begin by analyzing the relationship between the DMU and recurrent memory. Following this, we conduct an ablation study to investigate the impact of various hyperparameters employed in the DMU. Additionally, we introduce a simple yet effective thresholding scheme that can significantly reduce the computational cost of the proposed DMU. Furthermore, we extend the application of the delay line structure to popular LSTM and GRU models, showcasing its generalizability. Finally, we provide evidence that the DMU  enhances the predictive capability of the model. 

\subsection{Understand The Interplay Between Recurrent Memory and Delay Line}

To delve deeper into the relationship between the memory established by recurrent connections and the delayed lines, we incorporated the DMU into a one-layer Independently Recurrent Neural Network (Indrnn) comprising 512 neurons \cite{li2018independently}. The Indrnn architecture is defined as \( h_t = \sigma(W_{h}x_t + U_{h} \odot h_{t-1}) \), where \( U_{h} \) represents the self-recurrent weight vector that has the same dimensionality as the hidden state $h_{t-1}$, and \( \odot \) denotes the Hadamard product. Notably, in an IndRNN layer, each neuron operates independently and maintains its own memory state, thereby facilitating independent analysis. In this analysis, we use the SHD dataset and the loss function is derived from the output of the last time step. Thus, the network is trained to retain long-term memory. 

\begin{figure*}[]
\centering
\subfloat[]
{\includegraphics[width=0.45\textwidth]{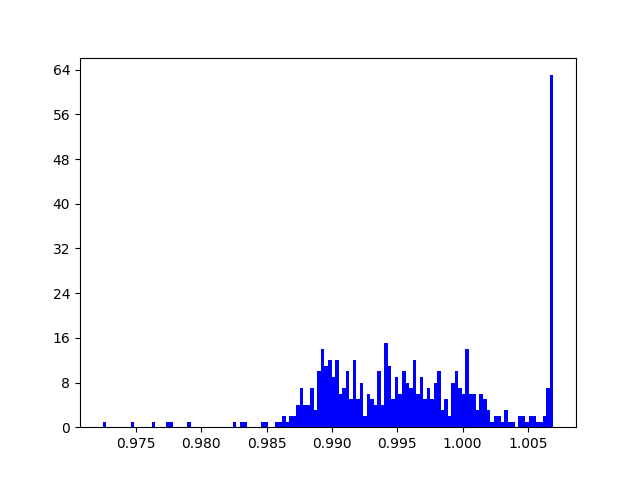}} 
\subfloat[]{\includegraphics[width=0.45\textwidth]{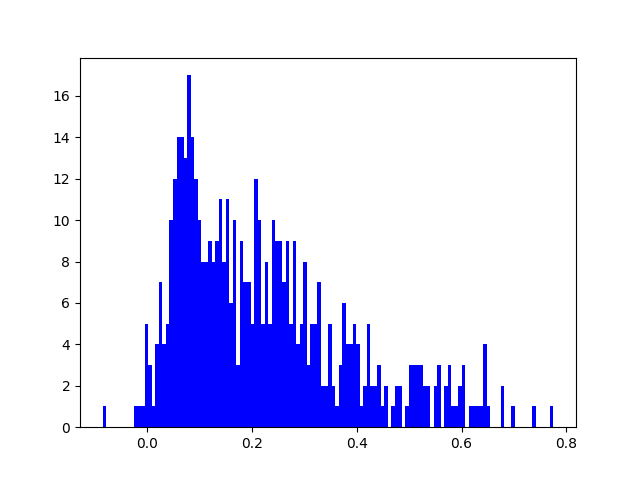}} 

\subfloat[]{\includegraphics[width=0.45\textwidth]{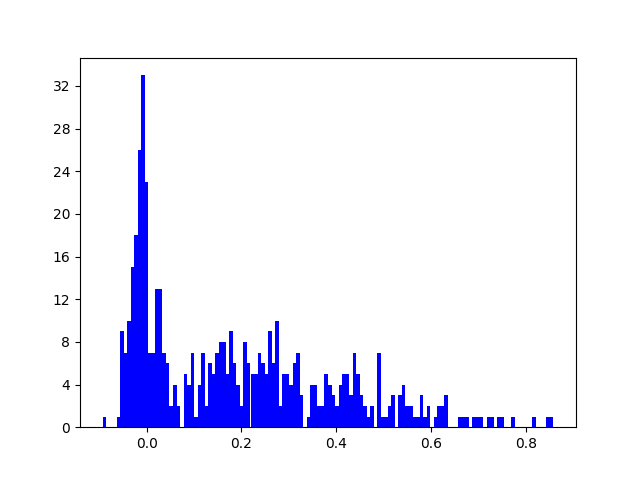}}
\subfloat[]{\includegraphics[width=0.45\textwidth]{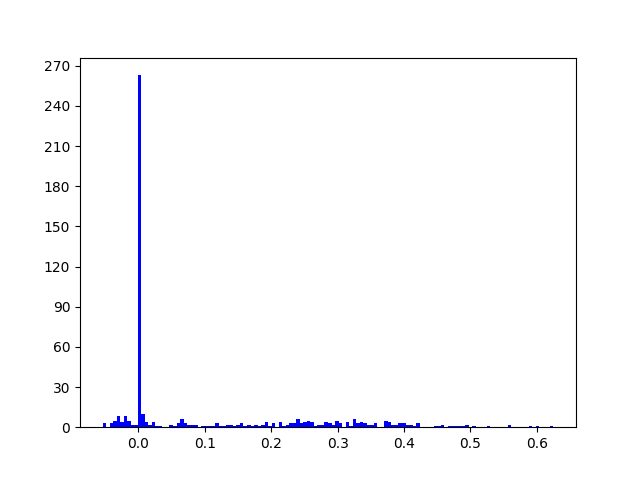}} 
\caption{Histograms of the learned recurrent weights on the  SHD dataset. (a) Indrnn.  (b) Indrnn+DMU (20 delays).  (c) Indrnn+DMU (60 delays).  (d) Indrnn+DMU (100 delays).  The x-axis represents the value of the recurrent weights and the y-axis represents the frequency. }
\vspace{0.2cm}
\label{shd_indrnn}
\end{figure*}

Fig. \ref{shd_indrnn} illustrates the histograms of the learned recurrent weights \( U_h \). While the conventional IndRNN layer achieves a test accuracy of 53\%, integrating IndRNN with DMU leads to a significant performance enhancement, with accuracy reaching 78\%. This notable improvement underscores the efficacy of the proposed DMU in enhancing temporal modeling. As shown in Fig. \ref{shd_indrnn}(a), a substantial portion of the weights in the vanilla Indrnn are close to 1, indicating these neurons are trained to retain long-term memory. In contrast, upon integration with DMU, the distribution of recurrent weights becomes more dispersive, suggesting a shift of the burden of modeling long-term dependencies to the delay line. Furthermore, the mean value of weights decreases with the increase in delay line length, attributed to the delay line's facilitation of more direct temporal credit assignment. Consequently, the model is encouraged to rely more on the delay line for establishing temporal dependencies.

\subsection{Effect of the Number of Delays and Delay Dilation Factors}
\label{effect_tau}
\vspace{-0mm}
\begin{figure*}[htb]
\centering
\subfloat[]{\includegraphics[width = 0.48\linewidth]{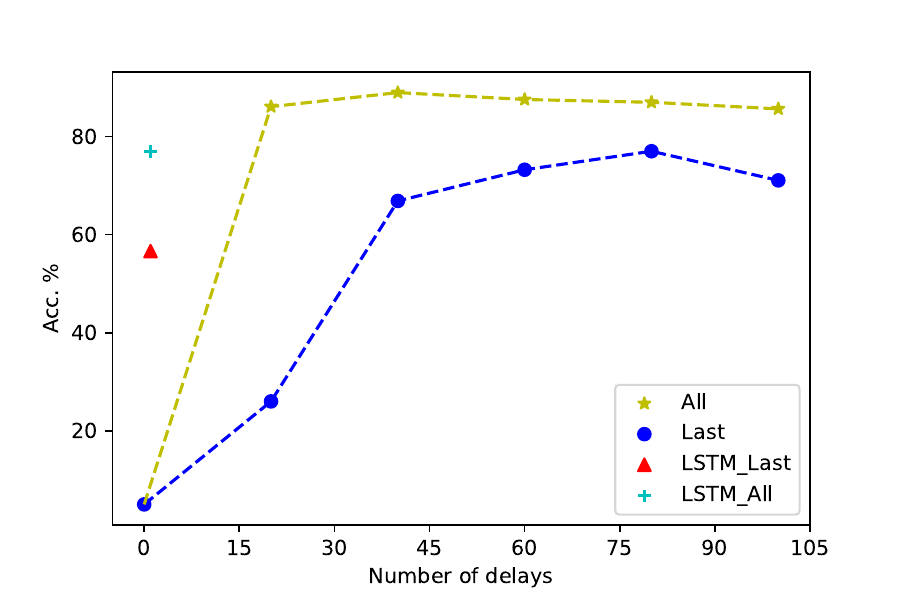}}  
\subfloat[]{\includegraphics[width = 0.48\linewidth]{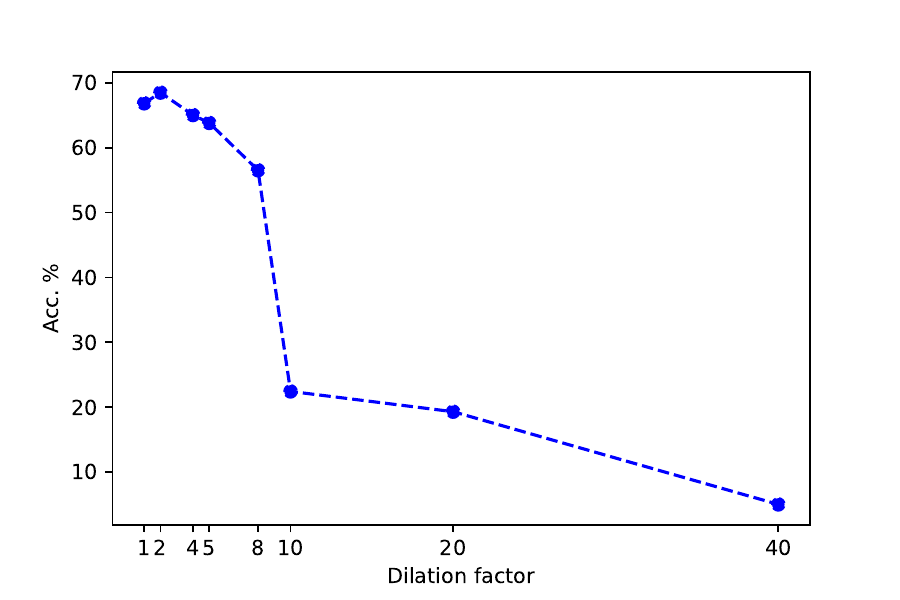}}
\caption{{\textbf{(a)} Comparison of performance for two decoding methods in DMU: Last time step loss (Last) and readout integrator (All). The X-axis represents the number of delays $n$ in a delay line and the Y-axis indicates classification accuracy. "LSTM\_Last" represents the LSTM coupled with the ``Last" decoding method, while ``LSTM\_All" stands for LSTM coupled with the ``All" decoding method. \textbf{(b)} The effect of the dilation factor $\tau$ on the classification accuracy. A fixed total delay of 40 has been used.}}

\vspace{0.2cm}
\label{dilation}
\end{figure*}
To investigate the impact of the number of delays $n$ and the dilation factor $\tau$ on temporal classification performance, we conducted an ablation study on the SHD dataset. The network architecture chosen for this study can be represented as 700-128-128-20, wherein the numbers indicate the number of neurons in each respective layer. Additionally, we compared two decoding methods: last-time step (Last) and readout integrator (All) \cite{yin2021accurate}. As depicted in Fig.~\ref{dilation}\textbf{(a)}, without utilizing the delay line ($n=0$), the classification accuracy remains at 5\%. This observation suggests that incorporating the delay line significantly enhances the capability of the vanilla RNN in sequential modeling. Furthermore, the `Last' decoding approach demonstrates improved performance with longer delay lines, indicating the importance of long-range temporal dependency for this task, as longer delay lines facilitate the establishment of such dependency. In contrast, the `All' method utilizes both the integrator and the delay lines to establish temporal dependencies. {Our experimental results suggest that incorporating longer delays generally improves performance, though it also leads to higher memory consumption. Once the delays surpass a certain threshold, the model's performance begins to saturate since the delay line has already fulfilled the necessary range of temporal dependencies as shown in both Fig.~\ref{dilation}\textbf{(a)} and Fig.~\ref{psmnist_1}. In practice, we have discovered that setting delay lengths between 30 and 80 time steps proves effective in achieving competitive results compared to gated RNNs across various tasks and datasets.}

{Figure \ref{dilation}\textbf{(b)} presents the results of the `Last' decoding method using different configurations of $\tau$. Our experimental results demonstrate that performance can be maintained at a promising level when $\tau$ is limited to a value of five or smaller. However, it is important to note that excessively high values beyond this threshold, which correspond to prolonged skip intervals, can result in the loss of fine-grained temporal information. This loss, in turn, has a detrimental effect on the model's performance. }

\subsection{Delay Gate Thresholding Scheme to Reduce Memory Usage}
\label{threshold_scheme}
\begin{figure*}[tbp]
\centering
\begin{minipage}[t]{0.5\textwidth}
\centering
\includegraphics[width=1\linewidth]{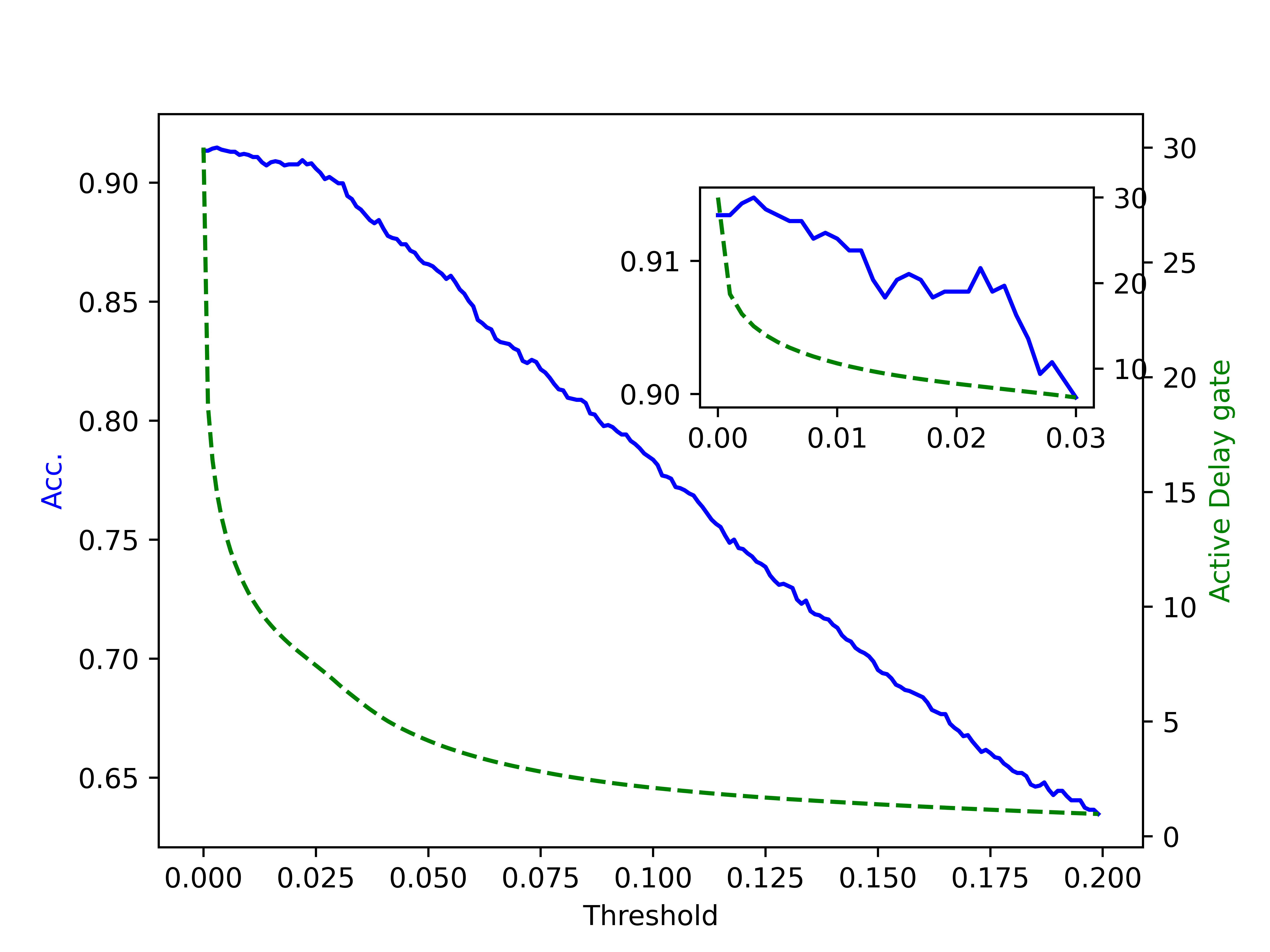}

\caption{Illustration of the impact of delay gate threshold on classification accuracy and the number of active delay gates. The total delay line length $n=30$. The left Y-axis corresponds to accuracy, while the right Y-axis represents the number of active delay gate.}
	\label{fig:example}
\end{minipage}
\begin{minipage}[t]{0.48\textwidth}
\centering
\includegraphics[width=1\linewidth]{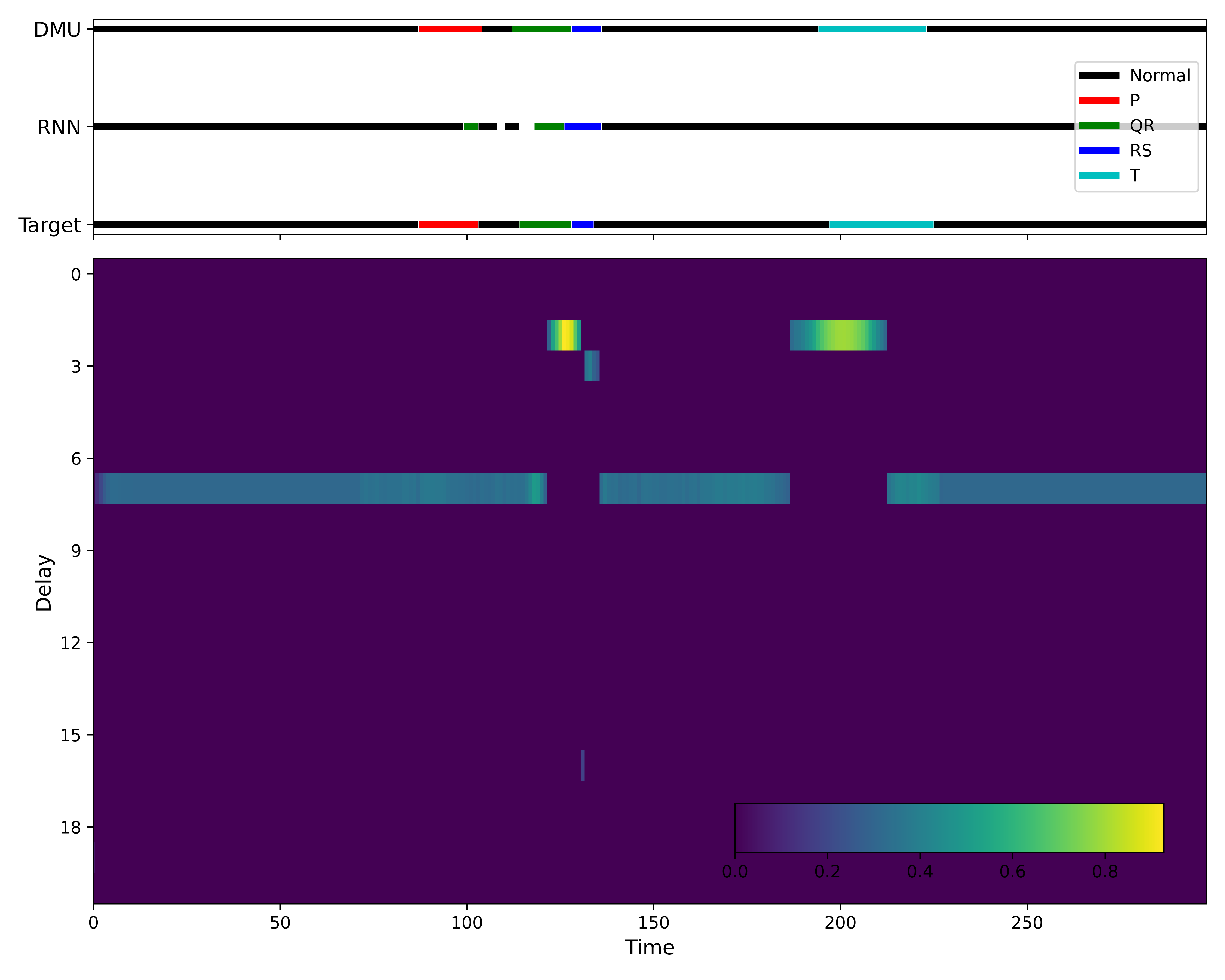}
\caption{Illustration of the delay gate values over time. A delay line with 21 delays has been used in this study. The top figure shows the output, presented from top to bottom: the target, the predicted label from the RNN model, and the predicted label from the DMU.}
\label{ecfclassification}
\end{minipage}
\end{figure*}

Here, we conduct a detailed analysis on how the proposed thresholding scheme affects both model performance and memory usage. In Fig. \ref{fig:example}, we present the results of applying various hard thresholds $\theta$ directly during inference for the SHD dataset. In the inset, it can be observed that the accuracy shows a slight improvement as the threshold value increases. This enhancement can be attributed to certain delay gates that unintentionally allow noise to pass through. By shutting these gates, the network effectively mitigates such noise interference, leading to improved performance. Interestingly, there is an initial significant reduction in computational overhead as the threshold increases. It should be noted that even when only 4 delay gates remain active, the proposed DMU achieves an accuracy surpassing 87\%, which represents SOTA performance. This highlights the effectiveness of the strategy, providing similar effect to the introduction of delay gate dilation. However, the proposed thresholding scheme represents a more dynamic and adaptable approach to achieving these improvements.

 \subsection{Empowering Gated RNNs with DMU's Delay Line Mechanism}
To further demonstrate the effectiveness of the DMU's delay line mechanism, we integrate it into well-established gated RNN architectures, namely GRU and LSTM. In this study, both models are configured to have a single recurrent layer consisting of 512 units. The results obtained on the SHD dataset, as presented in Table \ref{channel}, clearly illustrate the advantages of incorporating the delay lines. Specifically, with a modest increase of only 0.88\% in parameters, the LSTM model achieves a remarkable improvement of 15.99\% in accuracy on the SHD dataset. Similarly, the GRU model, when combined with the delay line mechanism, also exhibits substantial performance enhancements of nearly 10\%.

\begin{table}[ht]
	\caption{Comparison of LSTM and GRU model performance on the SHD dataset with and without incorporating delay lines. }
\small
	\centering
	\begin{tabular}{ccc}
		\cline{1-3}
  		\multicolumn{1}{c}{\bf Method}&
		\multicolumn{1}{c}{\bf Accuracy}&
	  \multicolumn{1}{c}{\bf Increased Parameters}
		\\ \hline
  	GRU &  $78.87\%$& - \\
      DMU-GRU ($n$ =  30) &   $\bf 88.43\%$ &1.17\%\\	
 	LSTM & 73.28\% & - \\
      DMU-LSTM  ($n$ = 30) &   $\bf 89.27\%$ &0.88\%\\
	  \hline 
	\end{tabular}
	\vspace{-0.0cm}

 	\label{channel}
 \end{table}

\subsection{DMU Enhance Model's Predictive Capability}
In contrast to conventional RNNs that summarize and store historical information in hidden states, the proposed DMU possesses the unique capability of directly projecting historical information into the future. This enables direct interaction and integration of temporally separated information, which greatly facilitates temporal pattern recognition. To demonstrate this capability, we conducted an experiment on the ECG dataset using a 1-layer RNN combined with a delay line of $n=21$. Accurately pinpointing the starting and ending points of the `RS' and `T' waves in this task is challenging due to their potentially biphasic nature \cite{exner2009noninvasive}. As illustrated in Fig. \ref{ecfclassification}, it is evident that the vanilla RNN struggles in accurately identifying the `RS' and `T' waves. In contrast, when equipped with the DMU, the network reliably projects information 7 time steps ahead, aligning with the temporal gap between the subsequent labeling points, which is approximately 7. Furthermore, the DMU dynamically toggles to a delay of 2 as the label is going to switch to `RS' and `T'. These sequential projections of information allow for the aggregation of relevant information, enabling the accurate classification of the waveform.

\section{Conclusion}
\label{conclusion}
In this study, we introduce DMU, a novel RNN architecture tailored for sequential modeling tasks. This model facilitates the direct establishment of temporal dependencies through its unique delay gate mechanism. Analysis of the distribution of learned memory and gate values suggests that the DMU enables the network to account for past events compared to conventional recurrent integration methods. Our experiments, spanning across audio processing (including classification and keyword detection), radar gesture recognition, ECG streaming classification, and Permuted sequential MNIST, have confirmed the effectiveness of the proposed methods. Notably, the DMU outperforms SOTA gated-RNN models in these tasks with significantly reduced model parameters. Furthermore, we propose two effective strategies to reduce the memory cost of DMU, including dilated delay line and delay gate thresholding schemes. {In the present study, the delay line operates at a uniform temporal scale governed by the dilation factor of the delay gate, which strikes a balance between computational efficiency and temporal modeling performance. However, real-world temporal signals exhibit intricate multiscale temporal dynamics. For instance, speech signals encompass various levels of structure, including phonemes, syllables, and words. Effectively and efficiently capturing such multiscale temporal patterns using a delay line remains an open question that necessitates further exploration in subsequent research efforts.}

\bibliographystyle{IEEEbib}
\bibliography{main}


\end{document}